\documentclass{article}

\usepackage{microtype}
\usepackage{graphicx}
\usepackage[dvipsnames]{xcolor}
\usepackage{subfigure}
\usepackage{booktabs} 
\usepackage{bm}
\usepackage{amsfonts,amsmath,amssymb,amsthm,mathtools}       
\usepackage{nicefrac}       
\usepackage{tikz}
\usepackage[inline, shortlabels]{enumitem}
\usepackage{thmtools}
\usepackage{thm-restate}
\usepackage{nicefrac}
\usepackage{multirow}
\usepackage{comment}

\graphicspath{ {./figures/} }

\usepackage{hyperref}

\usepackage[preprint]{tmlr}

\newcommand{\xhdr}[1]{\textbf{#1.}\;}

\newcommand{\bR}{\ensuremath \mathbb{R}}

\newcommand{\cF}{\ensuremath \mathcal{F}}

\newcommand{\cL}{\ensuremath \mathcal{L}}

\newcommand{\fstar}{\ensuremath f^{\star}}
\newcommand{\cFlin}{\ensuremath \mathcal{F}_{\mathrm{lin}}}
\newcommand{\Xtrue}{\ensuremath X^{\star}}

\DeclareMathOperator*{\argmin}{arg\,min}

\newcommand{\B}[1]{\bm{#1}}
\newcommand{\given}{\,|\,}

\newcommand{\dynotears}{\texttt{Dynotears}}

\newcommand{\reg}[1]{\ensuremath \|#1\|_{\mathrm{causal}}}
\newcommand{\approxreg}[2][]{\ensuremath \|#2\|_{\epsilon #1}}
\newcommand{\simplereg}[1]{\ensuremath \|#1\|_{\mathrm{simple}}}

\newcommand{\addref}[1]{{\color{Orange}[ref]}}

\theoremstyle{plain}

\theoremstyle{definition}

\usepackage[capitalize,noabbrev]{cleveref}

\theoremstyle{plain}

\theoremstyle{definition}

\theoremstyle{remark}

\title{Beyond Predictions in Neural ODEs:\\ Identification and Interventions}

\author{\name Hananeh Aliee \email hananeh.aliee@helmholtz-muenchen.de \\
      \addr Helmholtz Munich
      \AND
      \name Fabian Theis \email fabian.theis@helmholtz-muenchen.de \\
      \addr Helmholtz Munich
      \AND
      \name Niki Kilbertus \email niki.kilbertus@helmholtz-muenchen.de\\
      \addr Technical University of Munich \& Helmholtz Munich}
      
\begin{document}
\maketitle

\begin{abstract}
Spurred by tremendous success in pattern matching and prediction tasks, researchers increasingly resort to machine learning to aid original scientific discovery. Given large amounts of observational data about a system, can we uncover the rules that govern its evolution? 
Solving this task holds the great promise of fully understanding the causal interactions and being able to make reliable predictions about the system's behavior under interventions. 
We take a step towards such system identification for time-series data generated from systems of ordinary differential equations (ODEs) using flexible neural ODEs. 
Neural ODEs have proven successful in learning dynamical systems in terms of recovering observed trajectories.
However, their efficacy in learning ground truth dynamics and making predictions under unseen interventions are still underexplored.
We develop a simple regularization scheme for neural ODEs that helps in recovering the dynamics and causal structure from time-series data. 
Our results on a variety of (non)-linear first and second order systems as well as real data validate our method. We conclude by showing that we can also make accurate predictions under interventions on variables or the system itself.
\end{abstract}

\section{Introduction}
\label{sec:intro}

Many research areas increasingly embrace data-driven machine learning techniques not only for prediction, but also with the hope of leveraging data for original scientific discoveries.
We may formulate the core task of an ``automated scientist'' as follows:
\emph{Given observational data about a system, identify the underlying rules governing it!}
A key part of this quest is to determine how variables depend on each other.
Putting this question at its center, causality is a natural candidate to surface scientific insights from data.

Great attention has been given to ``static'' settings, where each observable under consideration is a random variable whose distribution is given as a deterministic function of other variables---its causal parents.
The structure of ``which variables listen to what other variables'' is typically encoded as a directed acyclic graph (DAG), giving rise to graphical structural causal models (SCM) \citep{pearl2009causality}.
SCMs have been successfully deployed both for inferring causal structure as well as estimating the strength of causal effects.
However, in numerous scientific fields, we are interested in systems that jointly evolve over time with dynamics governed by differential equations, which cannot be easily captured in a standard SCM.
In such interacting systems, the instantaneous derivative of a variable is a function of other variables (and their derivatives).
We can thus interpret the variables (or derivatives) that enter this function as causal parents~\citep{mooij2013ordinary}.
Unlike static SCMs, this accommodates cyclic dependencies and temporal co-evolution~\citep{bongers2016theoretical}.

The hope is that machine learning may sometimes be able to deduce true laws of nature purely from observational data, promising reliable predictions not only within the observed setting, but also under interventions.\footnote{An automated scientist would undoubtedly be more powerful if it can interact with the system and perform experiments.
By focusing on the purely observational setting, we avoid an ad-hoc specification of which experiments can be conducted and adhere to current settings, where algorithms are not granted direct access to real-world interventions.}
In this work we take a step towards system identification and causal structure inference from time-series data of a \emph{fully observed} system that is assumed to \emph{jointly evolve according to an ODE}.
We use scalable and flexible neural ODE estimators~\citep{chen2018neural} that allow for \emph{a-priori unknown non-linear interactions}. 
We start with noise-free observations in continuous time, but also provide results for additive observation noise, irregularly sampled time points, and real gene expression data.
Generally, recovering the ODE from a single observed solutions is an ill-posed problem.
However, existing theory suggests (informally) that for certain classes of ODEs ``most systems'' will be identifiable.
Motivated by these findings, we make the following contributions:
\begin{itemize}[leftmargin=*,topsep=0pt,itemsep=-1pt]
  \item We discuss potential regularizers to enforce sparsity in the number of causal interactions such that variables depend on few other variables, a common assumption in modern causal modeling~\cite{Scholkopf2021}.
  \item We develop a causal structure inference technique from time-series data called \emph{causal neural ODE} (\textbf{C-NODE}) combining flexible neural ODE estimators with suitable regularization techniques.
  C-NODE works for non-linear ODEs with cyclic causal structure and identifies the full ODE in certain cases. 
  Our results suggest that the proposed regularizer improves identifiability of the governing dynamics.
  \item We demonstrate the efficacy of C-NODE on a variety of low-dimensional \mbox{(non-)linear} first and second order systems, where it also makes accurate predictions under \emph{unseen interventions}.
  On simulated autonomous, linear, homogeneous systems we show that C-NODE scales to tens of variables. 
  \item Finally, C-NODE yields promising results for gene regulatory network inference on real, noisy, irregularly sampled single-cell RNA-seq data.
\end{itemize}

\noindent\textbf{Related work.}
There is a large body of work on the discovery of causal DAGs within the static SCM framework \citep{heinze2018causal,glymour2019review,vowels2021d}.
One key idea for causal discovery on time-series data is based on \emph{Granger causality}, where we attempt to forecast one time-series based on past values of others \citep{Granger1988}.
We review the basic ideas and contemporary methods in Section~\ref{sec:granger}.
Typically, these methods also return an acyclic directed interaction model, though feedback of the forms $X_t \to X_{t+1}$ or $X_t \to Y_{t+1}$ and $Y_t \to X_{t+1}$ is allowed.
Inferring Granger causality often relies on conditional independence tests \citep{Malinsky2018} or score-based methods \citep{Pamfil2020}.
Certain extensions of SCMs to cyclic dependence structures that retain large parts of the causal interpretation \citep{bongers2016theoretical} also allow for causal discovery of cyclic models \citep{lacerda2012discovering,hyttinen2012learning,mooij2011causal}.
The framing of our work differs from the above in that they cannot model evolutions of instantaneously interacting systems and only aim at the causal structure instead of the system specifics.

Another line of research has explored connections between (asymptotic) equilibria of differential equations and (extended) SCMs that preserve behavior under interventions \citep{mooij2013ordinary,bongers2018random,rubenstein2018deterministic,blom2020beyond}.
\citet{pfister2019learning} focus on recovering the causal dependence of a single target variable in a system of differential equations by leveraging data from multiple heterogeneous environments.
Following earlier work~\citep{dondelinger13a,raue2015data2dynamics,BENSON197997,Ballnus,Champion22445}, they consider mass-action kinetics only taking into account linear combinations of up to degree one interactions of the target variable's parents.
They enforce sparsity by only allowing a fixed number of such terms to be non-zero.
More broadly, parameter identifiability and estimation has been thoroughly investigated for discrete dynamical systems \citep{mcgoff2015statistical}, when the entire solution space is available \citep{grewal1976identifiability}, or time-lag is required with no instantaneous interactions~\citep{runge2018causal,ye2015distinguishing}.
Building on \citet{takens1981detecting}, the ``convergent cross mapping method'' \citep{sugihara2012detecting} overcomes separability assumptions of Granger causality and has successfully been extended to identify causal structure (and sometimes systems) for chaotic, strongly non-linear, or time-lagged systems among others \citep{ye2015distinguishing,Runge_2019,de2020latent}.
Current methods for ODE parameter estimation (with known parametric form) often deal with structural \citep{cobelli1980parameter} and practical (e.g., partial observability \citep{raue2009structural}) non-identifiability via empirical uncertainty analysis \citep{raue2015data2dynamics}.
For example, Sindy~\citep{brunton2016discovering}, a popular sparse regression method for identification of nonlinear systems, explicitly assumes sparsity in a set of candidate basis functions. This poses a limitation on the scalability of the method as the dictionary size grows combinatorially in the number of variables that are allowed to interact and nested non-linearities have to be added explicitly to the dictionary.
Hence, \citet{brunton2016discovering} acknowledge that Sindy fails to recover the dynamics already for a system of seven variables and that allowing for ``a broader function search space is an important
area of current and future work''.

In contrast to the works above, we assume neither a semantically meaningful pre-specified parametric form of the ODEs with a small set of parameters, nor the existence of equilibria.
We consider fully observed, non-delay ODEs (no time-lag, only instantaneous interactions) with observations from a single environment.
Our focus is on scalability to many variables with unknown non-linearities and leverage neural networks as flexible, yet efficiently learnable, function approximators.\footnote{Local minima are always a concern in high-dimensional non-convex optimization (training), but orthogonal to our work.}
Finally, we aim at fully identifying the ODE system, not only the causal structure, to be able to make predictions under interventions with a special focus on the multivariate case, i.e., systems of ODEs with sparse dependency structures.
Note that the causal structure is implied by the ODE system, hence ODE identification is strictly harder than causal structure identification in our setting.
In most applications, we consider the causal structure as an informative byproduct of our attempt at ODE system identification.

\citet{vorbach2021causal,massaroli2021dissecting,bellot2021graphical} aim at interpreting the inner workings of NODEs. 
Our main difference to~\citet{vorbach2021causal,massaroli2021dissecting} is that they do not go beyond predictive performance and disregard whether the true system has been learned. \citet{vorbach2021causal,bellot2021graphical} do not discuss the behavior of the system under interventions.

This research has potential applications in diverse fields including biology~\citep{pfister2019learning} (e.g., gene-regulatory network inference~\citep{btx194,QIU2020265}), robotics~\citep{Murray,kipf2018neural}, and economics~\citep{Zhang}. 

\section{Setup and Background}
\label{sec:setup}

Assume we observe the temporal evolution of $n$ real-valued variables $\Xtrue: [a, b] \to \bR^n$ on a continuous time interval $a < b$, such that $\Xtrue$ solves the system of ODEs\footnote{We use $\Xtrue$ for the real observed function and $X$ for a generic function. We refer to components $X_i$ as the observed variables of interest. Similarly, $f^{\star}$ is the ground truth system and $f$ a generic one.
Lower case letters denote observations at a fixed time, e.g., $x_0 = X(t=0)$.}
\begin{align}
  \dot{X} &= \fstar(X,t), \text{ for } \fstar \in \cF \text{ where }\label{eq:assumption} \\
  \cF &:= \{ f: \bR^n \times [a, b] \to \bR^n \given f \text{ uniformly Lipschitz-continuous in $X$ and continuous in $t$} \} \,. \nonumber
\end{align}
That is, we observe a single solution trajectory of some ODE (determined by) $f^{\star} \in \cF$.
\begin{equation}\label{eq:maingoal}
   \text{Our \textbf{main goal} is the following identification task: Given $\Xtrue$, identify } f^{\star} \in \cF.
\end{equation}

This is the inverse problem of ``solving an ODE'', for which the celebrated Picard-Lindel\"of theorem guarantees the existence and uniqueness of a solution of the \emph{initial value problem} (IVP) $\dot{X} = f(X, t), X(a) = x_0$ for all $f \in \cF$ and $x_0 \in \bR^n$ on an interval $t \in (a - \epsilon, a + \epsilon)$ for some $\epsilon > 0$.
We remark that higher-order ODEs, in particular second-order ODEs $\ddot{X} = f(X, \dot{X}, t)$, can be reduced to first-order systems via $U := (X, \dot{X}) \in \bR^{2n}$, $\dot{U} = (\dot{X}, \ddot{X}) = (U_2, f(U, t)$.\footnote{Second order systems require $X(a)$ and $\dot{X}(a)$ as initial values for a unique solution.
In practice, when only $\Xtrue$ is observed, we assume that we can infer $\dot{X}^{\star}(a)$, either from forward finite differences or during NODE training, see Section~\ref{sec:node}.
Any higher-order ODE can iteratively be reduced to a first order system.}
Hence, it suffices to continue our analysis for first-order systems.

\subsection{Causal Interpretation}

In SCMs causal relationships are typically described by directed parent-child relationships in a DAG, where the causes (parents) of a variable $X_i$ are denoted by $pa(X_i) \subset X$.
For ODEs an analogous relationship can be described by which variables ``enter into $f_i$''.
Formally, we define the \textbf{causal parents of $X_i$ in system $f$}, denoted by $pa_f(X_i)$, as follows: $X_j \in pa_f(X_i)$ if and only if there exist $x_1, \ldots, x_{j-1}, x_{j+1}, \ldots, x_n \in \bR$ such that $f_i(x_1, \ldots, x_{j-1}, \bullet, x_{j+1}, \ldots, x_n): \bR \to \bR$ is not constant.
This notion analogously extends to second and higher order equations by defining $pa_f(X_i)$ as the variables $X_j$ for which any (higher order) derivative of $X_j$ enters $f_i$.
Thereby, identifying $f^{\star}$ in eq.~\eqref{eq:maingoal} also yields the causal structure---it is an immediate yet informative byproduct.

One of the key advantages of causal models compared to merely predictive ones is that they enable us to make predictions about hypothetical interventions not in the training data.
In the ODE setting, different types of interventions can be conceived of.
We will focus on the following types of interventions.
 
\begin{itemize}[leftmargin=*,topsep=0pt,itemsep=-1pt]
    \item \textbf{Variable interventions:}
    For one or multiple $i \in \{1, \ldots, n\}$, we fix $X_i := c_i, f_i := 0$ and replace every occurrence of $X_i$ in the remaining $f_j$ with $c_i$ (for some constant(s) $c_i \in \bR$).
    We interpret these interventions as externally clamping certain variables to a fixed value.
    \item \textbf{System interventions:}
    We replace one or multiple $f_i$ with $\tilde{f}_i$.
    Here we can further distinguish between \emph{causality preserving} system interventions in which the causal parents remain unchanged, that is $pa_f(X_i) = {pa}_{\tilde{f}}(X_i)$ for all $i$, and others.
\end{itemize}

As an illustration, consider an ODE describing the positions of masses in a spring-mass system.
A variable intervention could amount to externally keeping one mass at a fixed point in space.
A system intervention could describe changing the stiffness of some of the springs.
We will analyze variable interventions in (non-linear) ODEs and system interventions primarily in linear settings with interpretable system parameters like in the chemical reaction example.

\subsection{(Neural) Ordinary Differential Equations}
\label{sec:node}

In neural ODEs (NODE) a machine learning model (often a neural network with parameters $\theta$) is used to learn the function $f_{\theta} \approx f$ from data \citep{chen2018neural}.
Starting from the initial observation $\Xtrue(a)$, an explicit iterative ODE solver is applied to predict $\Xtrue(t)$ for $t\in (a,b]$ using the current derivative estimates from $f_{\theta}$.
The parameters $\theta$ are then updated via backpropagation on the mean squared error between predictions and observations.
We mostly build on an augmented variant called SONODE that also works for second order systems and estimates the initial values $\dot{\Xtrue}(a)$ in an end-to-end fashion \citep{Norcliffe2020}.

Recently, NODEs have been extended to irregularly-sampled timesteps~\citep{Rubanova2019}, stochastic DEs~\citep{Li2020,Oganesyan2020}, partial DEs~\citep{Sun2019}, Bayesian NODEs~\citep{Dandekar2020}, and delay ODEs~\citep{Zhu2021}.
While these extensions could benefit our method, we use vanilla SONODE to disentangle the performance of our method from tuning the underlying NODE method.
As discussed extensively in the literature, NODEs can outperform traditional ODE parameter inference techniques in terms of reconstruction error, especially for non-linear $f$ \citep{chen2018neural,Dupont2019}.
A subsequent advantage over previous methods is that we need not pre-suppose a parameterization of $f$ in terms of a small set of semantically meaningful parameters.

Recently, a number of regularization techniques has been proposed for NODEs, where NODEs are viewed as infinite depth limits of residual neural networks (typically for classification) and there is no single true underlying dynamic law \citep{Finlay20, kelly2020easynode, Arnab20, pal2021opening, grathwohl2019ffjord}.
We emphasize that these existing techniques regularize the number of model evaluations to improve efficiency in learning one out of many possible dynamics yielding good performance on a downstream predictive task.
They are unrelated to our regularizer, which aims at identifying a single true underlying dynamical system from time series data.
Our regularizer targets neither the number of function evaluations, nor sparsity in the weights of the neural network directly (like pruning), but the number of causal dependencies.
Finally, another alternative to train neural networks that depend on few inputs, regularizing input gradients~\citep{Ross17,Ross17_2,Slavin2018TheNL} does not scale to high-dimensional regression tasks.

\subsection{Granger Causality}
\label{sec:granger}

Granger causality is a classic method for causal discovery in time series data that primarily exploits the directionality of time \citep{Granger1988}. 
Informally, a time series $X_i$ \emph{Granger causes} another time series $X_j$ if predicting $X_j$ becomes harder when excluding the values of $X_i$ from a universe of all time series.
Assuming that $X$ is stationary, multivariate Granger causality analysis usually fits a vector autoregressive model
\begin{equation}\label{eq:granger}
    X(t) = \sum_{\tau=0}^{k}W^{(\tau)} X(t-\tau) + E(t) ,
\end{equation}
where $E(t) \in \bR^n$ is a Gaussian random vector and $k$ is a pre-selected maximum time lag.
We seek to infer the $W^{(\tau)} \in \bR^{n \times n}$ from $X(t)$. 
In this setting, we call $X_i$ a Granger cause of $X_j$ if $|W^{(\tau)}_{i,j}| > 0$ for some $\tau \in \{0, \ldots, k\}$.
We need to ensure that $W^{(0)}$ encodes an acyclic dependence structure to avoid circular dependencies at the current time.
\citet{Pamfil2020} then estimate the parameters in eq.~\eqref{eq:granger} via
\begin{equation}\label{eq:dynotears}
  \min_{W} \Bigl\|X(t)-\sum_{\tau=0}^{k}W^{(\tau)} X(t-\tau)\Bigr\|_2 
  \! + \xi \|W^{(0)}\|_{1,1} + \rho \|W^{\backslash 0}\|_{1,1}\:,
\end{equation}
where $W = (W^{(\tau)})_{\tau=0}^k$, $W^{\backslash 0} = (W^{(\tau)})_{\tau=1}^k$, and $\|\cdot \|_{1,1}$ is the element-wise $\ell_1$ norm, which is used to encourage sparsity in the system. 
In addition, to ensure that the graph corresponding to $W^{(0)}$ interpreted as an adjacency matrix is acyclic, a smooth score encoding ``DAG-ness'' proposed by \citet{Zheng2018} is added with a separate regularization parameter.
While extensions to nonlinear cases exist \citep{Diks2016}, we primarily compare to \dynotears{} by \citet{Pamfil2020}---a choice motivated further in Appendix~\ref{app:disc}.

\section{Theoretical Considerations}
\label{sec:theory}

First, we note that our main goal in eq.~\eqref{eq:maingoal} is ill-posed.
A solution exists by assumption, but it may not be unique.\footnote{This violates one of the three Hadamard properties for well-posed problems. We use the word `solution' somewhat ambiguously and care must be taken not to confuse solutions to a given ODE or IVP (find $X$ given $f \in \cF$) and a solution of our main goal (finding $f^{\star} \in \cF$ given $\Xtrue$).}
We provide a simple example of two different autonomous, linear, homogeneous systems that have at least one solution in common in Appendix~\ref{app:proofs} (where we provide all proofs).\footnote{\emph{Autonomy} means that $f$ does not explicitly depend on time $f(X, t) = f(X)$. \emph{Linear} systems are ones where $f$ is linear in $X$, i.e., $f = A(t) X + b(t)$. \emph{Homogeneous} systems are linear systems in which $b(t) = 0$.}
This means that the underlying system is \emph{unidentifiable} from observational data.

Autonomous, linear, homogeneous systems are worthy a closer look:
\begin{equation}\label{eq:linsystems}
  \cFlin := \{ f(X) = A X \given A \in \bR^{n\times n} \} \subset \cF \,.
\end{equation}
First, they are common models for chemical reactions or oscillating physical systems.
Second, unlike for larger classes of ODEs, identifiability is reasonably well understood.
Within $\cFlin$ we can use $A$ and $f$ interchangeably.
For such systems, \citet{stanhope2014identifiability} developed beautiful graphical criteria for the identifiability of the system $A$ given $X^{\star}$, namely that the trajectory $X^{\star}$ is not confined to a proper subspace of $\bR^n$.
\citet{qiu2022identifiability} recently showed that an equivalent characterization is that $A$ has $n$ distinct eigenvalues, which implies that almost all $A \in \bR^{n \times n}$ are uniquely identifiable from $X^{\star}$ for almost all initial values $x_0\in \bR^n$.\footnote{Specifically, this holds with respect to the Lebesgue measure on $\bR^{n\times n}$ and $\bR^n$. The result still holds for probability measures from most common random matrix ensembles such as the Gaussian orthogonal, the Wishart, or the Ginibre ensembles.}
In this sense, all unidentifiable systems are non-generic and likely require some ``fine-tuning'' like our example in Appendix~\ref{app:proofs}, indicating that non-identifiability may not be a prevalent issue for the average case in practice.
While these results largely extend to affine linear systems \citep{duan2020identification}, little is known about identifiability in general non-linear systems \citep{miao2011identifiability}.
One may suspect non-identifiability to be a greater issue there, but highly non-linear or even chaotic systems are sometimes known to be identifiable \citep{takens1981detecting,sugihara2012detecting,de2020latent}.


While these results are encouraging, we are particularly interested in ``simple'' interactions, which we capture by sparsity.
That is, we assume that in natural systems each variable depends on only few other variables as causal parents \citep{Scholkopf2021}.
Counting the number of parent-child relationships in a system $f$ as $\reg{f} := \sum_{i=1}^n |pa_f(X_i)|$, we are thus interested only in possible ground truths $f^{\star}$ with small $\reg{f^{\star}}$ (compared to $n^2$).
For $\cFlin$, this amounts to matrices $A \in \bR^{n\times n}$ with at most $k$ non-zero entries.
To the best of our knowledge, it remains an open problem whether among such sparse matrices still almost all of them are identifiable from $X^{\star}$ (for almost all initial conditions $x_0$).
Since sparse matrices are more likely to have repeated eigenvalues, the existing theory for dense matrices does not carry over to our setting.
Hence, there is a gap between predictive performance (reconstruction error of $X$) and \emph{identifying the governing system}, which is required to make \emph{predictions under interventions}.
NODEs perform well in terms of predictive performance, but theoretically they may do so by learning the ``false'' system.
This is a key motivation for our empirical analysis in this work.
In the following, we develop a method to identify such sparse systems from a single observed solution trajectory via regularization and assess identifiability empirically not only in the linear, but also non-linear case.


We first formulate our \textbf{regularized goal}: Given $\Xtrue$, find $f \in \cF$ such that $\Xtrue$ solves $\dot{X} = f(X, t)$ and $f \in \argmin_{g \in \cF} \|g\|$ for some measure of complexity $\|\cdot \|: \cF \to \bR_{\ge 0}$.
Without knowing $f^{\star}$ a priori, $\reg{\cdot}$ is difficult to enforce as a complexity measure in practice.
Instead, we approximate the requirement of ``not being constant w.r.t.\ to an argument'' in our definition of causal parents via the following regularizer
\begin{equation}\label{eq:regularizer}
  \approxreg{f} := \sum_{i,j=1}^n \B{1}\{ \|\partial_j f_i\|_2 > \epsilon \}\quad  \text{ for some } \epsilon \ge 0,
\end{equation}
where $\|\cdot \|_2$ is the $L^2$ norm on the Hilbert space of square-integrable real functions (with respect to the Lebesgue measure).
For $A \in \cFlin$ this captures sparsity in the common sense $\approxreg[=0]{A} = \sum_{i,j=1}^n \B{1}\{A_{ij} \ne 0\} = \reg{A}$.\footnote{$\approxreg[=0]{A}$ and therefore also $\reg{A}$ is not a norm; it violates the triangle-inequality.}
There are still two hurdles to implementing $\approxreg{f}$ in practice.
(a) For the full non-linear case $f \in \cF$ the $L^2$-norms of partial derivatives are difficult to evaluate efficiently and accurately.
(b) Even for $A \in \cFlin$, $\approxreg{A}$ is not differentiable.
For the linear case, non-differentiability of $\approxreg{A}$ is typically overcome by enforcing sparsity via an entry-wise $\ell_1$ norm $\|A\|_{1,1} = \sum_{i,j=1}^n |A_{ij}|$ as a penalty term, like in eq.~\eqref{eq:dynotears}.
While this covers the linear case, in section~\ref{sec:method} we develop a technique to partially overcome problem (a) in the non-linear case, by reformulating $\approxreg{f}$ as an entry-wise $\ell_1$ norm for a matrix derived from the parameters $\theta$ of the neural network $f_{\theta}$ approximating $f$.



\xhdr{Remarks}
In the realm of neural ODEs one may be tempted to enforce sparsity in the neural network parameters $\theta$ directly.
While this can be a sensible regularization scheme to improve generalization~\citep{liebenwein2021sparse}, it does not directly translate into interpretable properties of the ODE $f_{\theta}$.
For fully connected neural networks even a sparse $\theta$ typically leads to dense input-output connections such that $\approxreg{f} = n^2$. 

Alternatively, one may train a separate neural network $f_{i,\theta_i}: \bR \times \bR^n \to \bR$ with parameters $\theta_i$ for each component $f_i$.
Stacking the outputs of all $f_{i,\theta_i}$ we can then train each network separately from the same training signal.
For such a parallel setup we can enforce sparsity via $\approxreg{f_i}^{\mathrm{single}} := \sum_{j=1}^n \B{1}\{ \|\partial_j f_i\|_2 > \epsilon \}$ in the first layer parameters of each component neural network separately to ``zero out'' certain inputs entirely \citep{bellot2021graphical}.
However, this differs from our sparsity measure $\approxreg{f}$ in eq.~\eqref{eq:regularizer}.
The latter allows large $\approxreg{f_i}^{\mathrm{single}}$ for \emph{some} components $f_i$ \emph{as long as the entire system is sparse}.
The parallel approach did not perform better in empirical experiments, but is computationally more expensive.


\section{Method}\label{sec:method}

\xhdr{Practical regularization}
Let us write out $f_{\theta}$ explicitly as a fully connected neural network with $L$ hidden layers parameterized by $\theta := (W^l, b^l)_{l=1}^{L+1}$, with $l$-th layer weights $W^l$ and biases $b^l$
\begin{equation}
    f_{\theta}(Y) = W^{L+1} \sigma(\ldots \sigma( W^2 \sigma(W^1 Y + b^1) + b^2) \ldots) \:, 
\end{equation}
with element-wise non-linear activations $\sigma$. 
With this parameterization we now approximate our desired regularization $\reg{f_{\theta}}$ in terms of $\theta$.
A major drawback of the natural candidate $\approxreg{f_{\theta}}$ from eq.~\eqref{eq:regularizer} is that it is piece-wise constant in $\theta$---an obstacle to gradient-based optimization.
Instead, we aim at replacing $\approxreg{f}$ with a differentiable surrogate.
Ideally, we would like to use $\ell_1$ regularization on the strengths of all input to output connections $j \to i$ in $f_{\theta}$.
\begin{itemize}[leftmargin=*,topsep=0pt,itemsep=-1pt]
    \item \textbf{The linear case.} Recall that for $\cFlin$ it suffices to choose linear activation functions (specifically, $\sigma(x) = x$) and $b^l = 0$, such that $f_{\theta}(Y) = A Y$ for some $A = W^{L+1}\cdot \ldots \cdot W^1$.\footnote{When biases are non-zero, we are in the realm of inhomogeneous, linear, autonomous systems.}
    Hence, for $f_{\theta} \in \cFlin$, we can directly implement a continuous $\ell_1$ surrogate of the desired regularizer in terms of $\theta$
    \begin{equation}\label{eq:practicalregularizer}
      \simplereg{\theta}^{\mathrm{lin}} := \|A\|_{1,1} = \|W^{L+1} \cdot \ldots \cdot W^1\|_{1,1} \:.
    \end{equation}
    We then have $X_j \in pa_{f_{\theta}}(X_i)$ if and only if $A_{ij} \ne 0$.
    When restricting ourselves to $\cFlin$, using $\sigma(x) = x, b^l = 0$ together with the above sparsity constraint is thus a viable and theoretically sound method.
    In this case we infer $A$ directly from $\theta$, i.e., we identify $f$, from which the causal structure follows.
    
    \item \textbf{The non-linear case.} In practice, we do not know whether $f^{\star}\in \cFlin$ a priori.
    Thus we remain open to the possibility of non-linear $f^{\star}$ via non-linear activation functions $\sigma$.
    Then $\simplereg{\theta}^{\mathrm{lin}}$ is not equivalent to $\reg{f_{\theta}}$ anymore, because we may have $X_j \not \in pa_{f_{\theta}}(X_i)$ despite $A_{ij} \ne 0$.
    In this case, we use the absolute weights for the regularizer: 
    \begin{equation}\label{eq:practicalregularizer:nonlinear}
    \simplereg{\theta}^{\mathrm{non-lin}}:= \||W^{L+1}| \cdot \ldots \cdot |W^1|\|_{1,1} \:.
    \end{equation}
    While we can still have $[|W^{L+1}| \cdot \ldots \cdot |W^1|]_{i,j} \ne 0$ even though $X_j \notin pa_{f_{\theta}}(X_i)$, $[|W^{L+1}| \cdot \ldots \cdot |W^1|]_{i,j}=0$ always implies $X_j \notin pa_{f_{\theta}}(X_i)$ for $b^l = 0$.
    Hence, using $\simplereg{\theta}^{\mathrm{non-lin}}$ as a regularizer aims at minimizing an upper bound of $\reg{f_{\theta}}$.
    We show empirically that enforcing this upper bound of the desired regularizer serves as an effective inductive bias to enforce sparsity in the causal connections.
\end{itemize}
From now on, we will drop the superscript of $\simplereg{\theta}$ when it is clear from context.

\xhdr{Causal structure inference}
While we could read off the causal structure directly from $\theta$ via $A$ in the linear case, for non-linear $f_{\theta}$ we can validate our results via partial derivatives ${\partial_j (f_{\theta})_i}$ over time, showing how each $\dot{X}_i$ depends on each $X_j$.
Following the reasoning of the regularizer $\approxreg{f_{\theta}}$ in eq.~\eqref{eq:regularizer}, we then reconstruct the causal relationships via $X_j \in pa_{f_{\theta}}(X_i)$ if and only if $\sum_{k=1}^{N}|{\partial_j f_{\theta,i}(t_k)}| > \epsilon$ for the $N$ observations at times $a=t_1 < \ldots < t_N = b$ and some threshold $\epsilon > 0$.\footnote{This is but one simple approach to estimate practically whether $\|\partial_j f_{\theta, i}\|_2 \ne 0$, i.e., whether the $i$-th output of the neural network $f_{\theta}$ depends on the $j$-th input. While a host of more sophisticated methods may be used here, we found this simple approach sufficient in our experiments.}
Thus we can still infer the causal structure in non-linear cases where evaluating whether our method identified the correct $f^{\star}$ is challenging (as we cannot compare parameters directly, but would have to compare a neural network to an analytically known function in symbolic form).
The choice of $\epsilon$ is sensitive to the scale of the data, which we account for by normalizing data before training.
Empirically we did not observe strong dependence of the inferred causal structure on the choice of $\epsilon$ for normalized data.
A simpler method is available to determine the \emph{absence} of causal dependencies in the non-linear setting: if the entry $[|W^{L+1}| \cdot \ldots \cdot |W^1|]_{i,j}$ is (close to) zero, then $X_j \notin pa_{f_{\theta}}(X_i)$.

\begin{figure}
\vspace{-1cm}
  \centering
  \includegraphics[width=0.65\columnwidth]{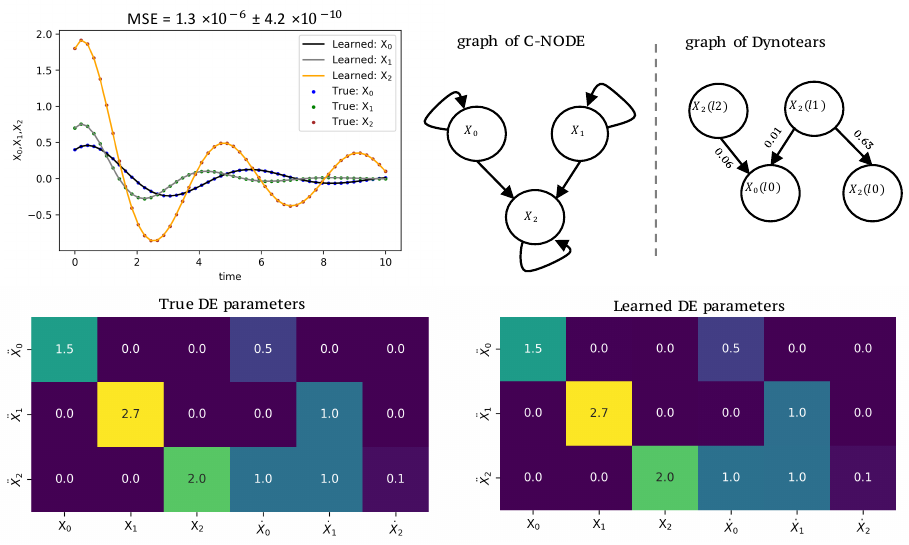}
  \caption{Exemplary second-order linear system.}\label{fig:linear}
 \vspace{-0.4cm}
\end{figure}

\xhdr{Summary}
C-NODE adds a practical, differentiable regularizer $\lambda \simplereg{\theta}$ with a tuneable regularization parameter $\lambda$ to the NODE loss to recover sparse dynamics corresponding to our regularized goal.
Our regularizer captures $\reg{f}$ perfectly in the linear case and enforces minimization of an upper bound in the non-linear case.
We also devised a method to recover the causal structure from linear and non-linear $f_{\theta}$.
In Section~\ref{sec:experiments} we show empirically that potentially remaining theoretical unidentifiability does not practically impede C-NODE from recovering $f^{\star}$, allowing for accurate predictions under variable and system interventions.
In Appendix~\ref{app:extensions} we discuss extensions for latent dynamics and data from heterogeneous environments.

\section{Experiments}
\label{sec:experiments}

\begin{figure}
\vspace{-1cm}
  \centering
  \includegraphics[width=0.4\columnwidth]{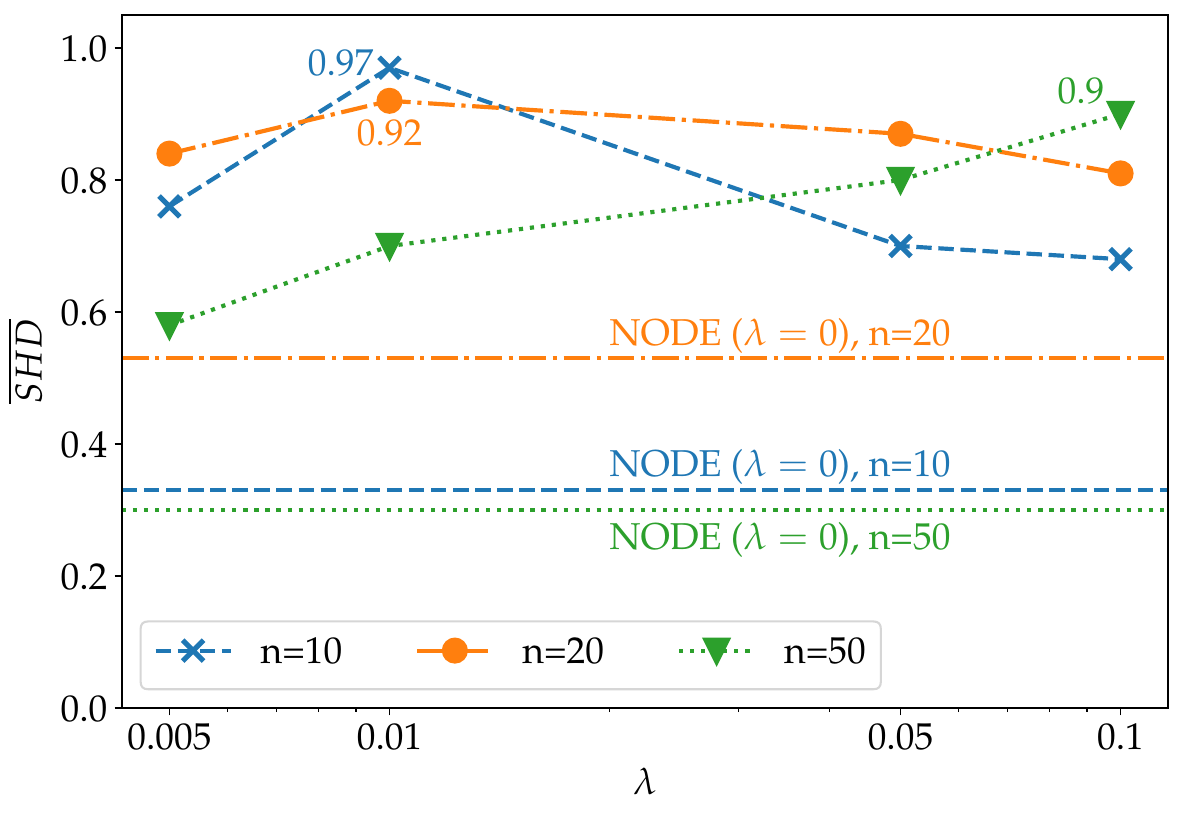}
  \caption{Dependence on the sparsity parameter $\lambda$. Dashed lines labeled $\lambda=0$ refer to vanilla NODE.}\label{fig:lambda}
\end{figure}

We now illustrate the robustness of our method in several case studies. 
The general principles readily extend to more complex model classes.
Among several methods developed for causal inference from time-series data based on Granger causality \citep{Tank_2021,hyvarinen10a,Runge_2019,Amornbunchornvej_2019}, we compare to \dynotears{} \citep{Pamfil2020}, because it outperforms most competing methods in their evaluation.
For system identification, we also compare to GroupLasso \citep{bellot2021graphical} and PySINDy \citep{brunton2016discovering,desilva2020}.
Details on all parameter settings, evaluation, and implementation choices are provided in Appendix~\ref{app:implementation}.

\xhdr{Linear ODEs}
We first study second-order, homogeneous, autonomous, linear ODEs
\begin{equation}\label{eq:linear}
    \ddot{X} = W_1\dot{X} + W_2X .
\end{equation}
We begin with $n=3$ and randomly chosen true weight matrices $W^{\star}_1, W^{\star}_2$ from which we generate $\Xtrue$ using a standard ODE solver, see Appendix~\ref{app:implementation}.
Figure~\ref{fig:linear} shows that our method not only accurately predicts $\Xtrue$, but it also identifies $W_1^{\star}, W_2^{\star}$ within a maximum absolute difference of $0.018$.
Thus the causal graph is also inferred correctly.
The poor performance of \dynotears{} may be due to cyclic dependencies in $W_1^{\star}, W_2^{\star}$.

\begin{table}[t]
 \centering
 \caption{Experimental results using synthetic datasets with $n \in \{10, 20, 50\}$, varying noise level $\sigma$, and sampling irregularity (irr). SHD is the structural Hamming distance and $\overline{\text{SHD}}$ = 1 - SHD.}\label{tab:results}
 {
\scriptsize
 \begin{tabular}{c c c c c c c c c c c c c c} \toprule
 ~ & ~ & \multicolumn{3}{c}{irr = 0.0} & \multicolumn{3}{c}{irr = 0.2} & \multicolumn{3}{c}{irr = 0.5} & \multicolumn{3}{c}{irr = 0.7}\\
 \cmidrule(lr){3-5}\cmidrule(lr){6-8}\cmidrule(lr){9-11}\cmidrule(lr){12-14}
 $\sigma$ & dim & $\overline{\text{SHD}}$ & TPR & TNR & $\overline{\text{SHD}}$ & TPR & TNR & $\overline{\text{SHD}}$ & TPR & TNR & $\overline{\text{SHD}}$ & TPR & TNR\\
 \midrule
 \multirow{3}{*}{0.0} & 10 & 0.97 & 0.95 & 0.98 & 0.98 &0.97 &0.98 &0.93 &0.97 &0.91 &0.92 &0.93 &0.91\\ 
 ~ & 20 & 0.92 & 0.82 & 0.97 & 0.84 & 0.72 & 0.88 & 0.85 &0.74 &0.88 &0.86 &0.75 &0.89\\ 
 ~ &50 & 0.90 & 0.71 & 0.92 & 0.85 & 0.67 & 0.87 & 0.92 &0.69 &0.96 & 0.93 & 0.70 & 0.96\\
 \midrule
 \multirow{3}{*}{0.05} & 10 & 0.88 & 0.81 & 0.92 & 0.87 & 0.80 & 0.92 &0.86 & 0.84 & 0.88 & 0.67 & 0.48 & 0.87\\
 ~ & 20 & 0.86 & 0.78 & 0.89 & 0.84 & 0.72 & 0.88 & 0.72 & 0.59 & 0.75 &0.69 & 0.53 & 0.73 \\
 ~ & 50 & 0.91 & 0.64 & 0.94 & 0.90 & 0.69 & 0.93 & 0.90 & 0.67 & 0.93 & 0.89 & 0.65 & 0.93\\
 \midrule
 \multirow{3}{*}{0.1} &10 & 0.78 & 0.68 & 0.74 & 0.71 & 0.68 & 0.71 & 0.65 & 0.77 & 0.58 & 0.50 & 0.60 & 0.51\\ 
 ~ &20 & 0.82 & 0.79 & 0.82 & 0.79 & 0.74 & 0.80 &0.68 & 0.53 & 0.74 & 0.61 & 0.48 & 0.65\\ 
 ~ &50 & 0.86 & 0.65 & 0.90 & 0.89 & 0.59 & 0.93 & 0.87 & 0.61 & 0.91 & 0.86 & 0.68 & 0.89  \\
 \bottomrule
\end{tabular}}
\end{table}

We extend these results to study the \emph{scalability} of our method and its performance when the observations are \emph{irregularly} sampled (a fixed fraction of observations is dropped uniformly at random) with \emph{measurement noise} (additive zero-mean Gaussian noise with standard deviation $\sigma$).
The data generation specifics for three synthetic datasets with 10, 20, and 50 variables are described in Appendix~\ref{app:implementation}.
We evaluate the inferred causal graph using the structural hamming distance (SHD)~\citep{Lachapelle2019} for the fraction of wrongly predicted edges, the true positive rate (TPR), and the true negative rate (TNR).
The results in Table~\ref{tab:results} show that C-NODE performs well for non-noisy data ($\sigma=0$) and is robust to randomly removing samples from the observation. 
Accuracy drops with increasing noise levels, which is further exacerbated by sampling irregularities, suggesting improved robustness to observation noise as an interesting direction for future work.
In Figure~\ref{fig:lambda} we show the dependence of $\overline{\text{SHD}} = 1 - \text{SHD}$ on the regularization parameter $\lambda$ for high-dimensional systems following eq.~\eqref{eq:linear}.
C-NODE clearly outperforms vanilla NODE for causal structure inference with only moderate sensitivity to $\lambda$ even though they achieve similar predictive accuracy.
This indicates that for sparse systems unidentifiability may indeed be a problem (as hypothesized in section~\ref{sec:theory}) in that the solution trajectory can be perfectly reconstructed also by ``false'' dense systems.
This provides evidence that regularization is indeed crucial for recovering the correct sparse ground truth dynamics.
In Figures~\ref{fig:synthetic:adjacency} and \ref{fig:SHD}, we provide further results and show that C-NODE also outperforms both vanilla NODE~\cite{chen2018neural},  GroupLasso~\citep{bellot2021graphical} and PySINDy~\citep{desilva2020}.

A concrete application example for a common (synthetic) chemical reaction network of transcriptional gene dynamics is also provided in Appendix~\ref{app:chemical}.

\begin{figure}
\vspace{-1cm}
    \centering
    \includegraphics[width=0.7\columnwidth]{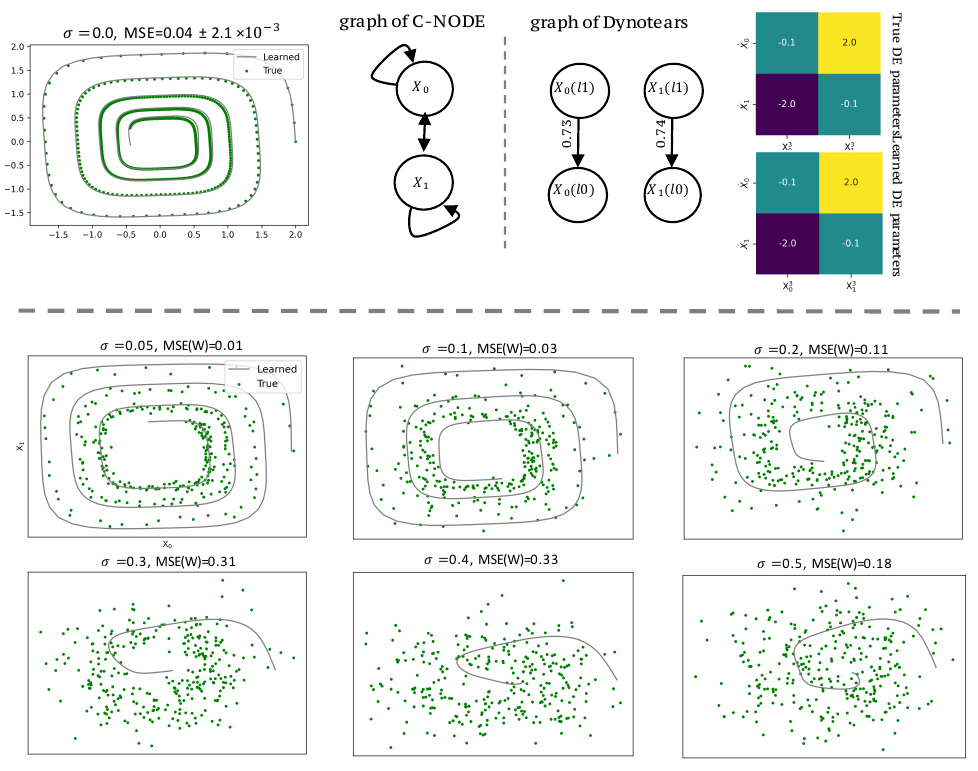}
    \caption{Results for the spiral ODE.}\label{fig:spiral}
\end{figure}

\xhdr{Spiral ODEs}
The spiral ODE model is given by
\begin{equation}
    \dot{X}_0 = -\alpha {X_0}^3 + \beta {X_1}^3, \qquad
    \dot{X}_1 = -\beta {X_0}^3 + \alpha {X_1}^3
\end{equation}
and features cyclic dependencies and self-loops.
We follow the parameterization in~\citet{chen2018neural}. 
While \dynotears{} fails to estimate the cyclic causal graph, Figure~\ref{fig:spiral} shows that C-NODE infers the actual ODE parameters $\alpha, \beta$ and thus the causal structure correctly. 
Again, Figure~\ref{fig:spiral} illustrates that predictive performance slowly degrades as observation noise levels increase raising the mean-squared error (MSE) of the inferred adjacency matrix substantially for higher variance noise. 
However, the deduced causal structure remains correct. 

\begin{figure}
\vspace{-1cm}
    \centering
    \includegraphics[width=0.6\columnwidth]{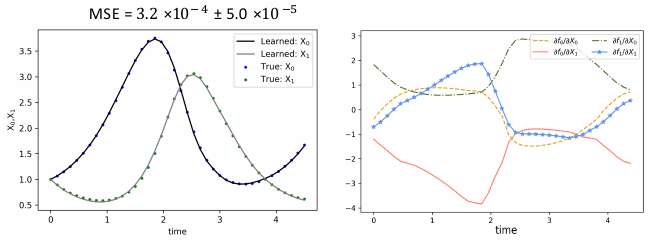}
    \caption{Results for the Lotka-Volterra example.\label{fig:LV}}
\end{figure}

\xhdr{Lotka-Volterra ODEs}
The Lotka-Volterra predator-prey model is given by the non-linear system
\begin{equation}\label{eq:LV0}
    \dot{X}_0 = -\alpha X_0 - \beta X_0 X_1, \quad
    \dot{X}_1 = -\delta X_1 + \gamma X_0 X_1.
\end{equation}
We use the same parameters as~\citet{Dandekar2020} and ReLU activations for non-linearity.
Figure~\ref{fig:LV} shows the excellent predictive performance of C-NODE (left).
Because of the non-linearity, we resort to our non-linear causal structure inference method and show the partial derivatives $\partial_j{f_{\theta,i}}$ for the learned $f_{\theta}$ in Figure~\ref{fig:LV} (right).
For example, from eq.~\eqref{eq:LV0} we know that $\partial \dot{X}_0 / \partial X_1 = - \beta X_1$ and indeed $\partial_1 f_{\theta,0}$ in Figure~\ref{fig:LV} (right) resembles $X_1$ in Figure~\ref{fig:LV} (left) up to rescaling and a constant offset.
Similarly, the remaining dependencies estimated from $f_{\theta}$ strongly correlate with the true dependencies encoded in eq.~\eqref{eq:LV0}, giving us confidence that $f_{\theta}$ has indeed correctly identified $f^{\star}$.

\begin{figure}
\centering
\includegraphics[width=0.6\columnwidth]{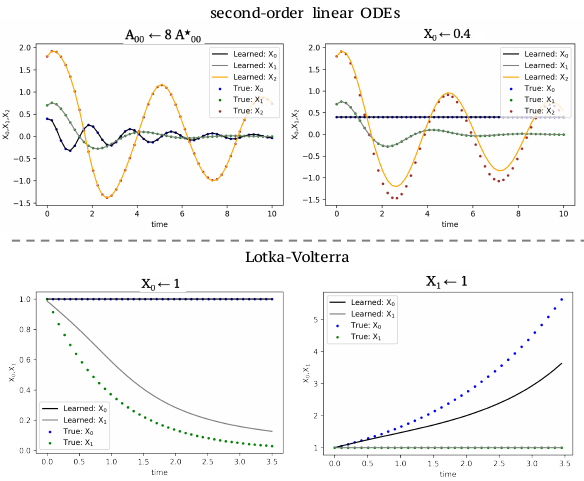}
\caption{Predictions under interventions.\label{fig:intervention}}
\vspace{-0.4cm}
\end{figure}

\xhdr{Interventions}
To back up this claim, we assess whether we can predict the behavior of systems under interventions.
We consider C-NODEs trained on observational data (without interventions) from Figures~\ref{fig:linear} (simple linear system) and \ref{fig:LV} (Lotka-Volterra).
We apply two types of interventions: (1) a system intervention replacing one entry of $A^{\star}$ via $\tilde{A}_{00} := 8 A^{\star}_{00}$ (for example, a temperature change that increases some reaction rate eightfold) in the linear setting, and (2) variable interventions $X_0 := 0.4$ in the linear as well as $X_0 := 1$ and $X_1 := 1$ in the non-linear Lotka-Volterra setting (for example, keeping the number of predators fixed via culling quotas and reintroduction).
Figure~\ref{fig:intervention} (top left) shows that C-NODE successfully predicts the linear system's evolution under the system intervention.
For the variable intervention in the linear setting (top right), $X_1$ correctly remains unaffected, while the new behavior of $X_2$ is predicted accurately.

In the Lotka-Volterra example, both variable interventions impact the other variable.
Fixing either the predator or prey population should lead to an exponential increase or decay of the other, depending on whether the fixed levels can support higher reproduction than mortality.
Figure~\ref{fig:intervention} (bottom row) shows that our method correctly predicts an exponential decay (increase) of $X_0$ ($X_1$) for fixed $X_1 := 1$ ($X_0 := 1$) respectively.
The quantitative differences between predicted and true values stem from small inaccuracies in the predicted parameters which amplify exponentially to seemingly large quantitative differences.

\xhdr{Real single-cell RNA-seq data}
Finally, we apply C-NODE to learn gene-gene interactions.
Gene (feature) interactions, also known as causal dependencies between genes, are often represented as a gene regulatory network (GRN) where nodes correspond to genes and directed edges indicate regulatory (or causal) interactions between genes.
GRN inference from observations is known to be an exceptionally difficult task~\cite{Perkel}. 
The inferred GRN is expected to be sparse as regulatory genes known as transcription factors do not individually target all genes. 
Therefore, sparsity in the number of interactions is essential. 

\begin{figure}[t]
\vspace{-1cm}
\centering
\includegraphics[width=0.9\columnwidth]{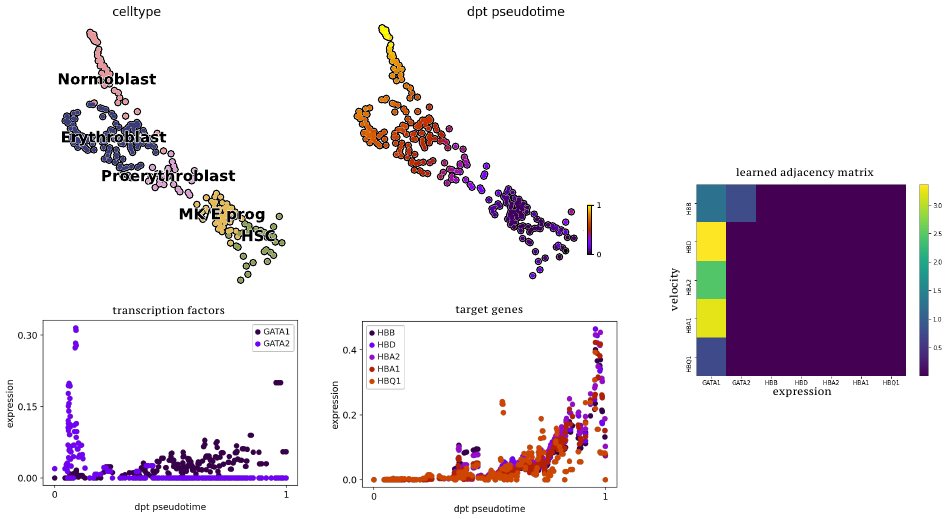}
\caption{Gene regulatory network inference results using human bone marrow data with 7 genes.}\label{fig:GRN_human}
\vspace{-0.4cm}
\end{figure}

In this experiment, we first explore how pruning improves GRN inference from human hematopoiesis single-cell multiomics data~\cite{luecken2021a}
(GEO accession code: GSE194122). 
We select a branch of data in which hematopoiesis stem cells (HSCs) differentiate into Erythroid cells with a total of 280 cells (or samples).
The count matrix is normalized to one million counts per cell.
Figure~\ref{fig:GRN_human} (top row) shows a UMAP representation \citep{McInnes2018} of the data where each point corresponds to a cell colored by cell type (left) and an inferred continuous pseudotime (right).
This pseudotime aims at identifying how far a cell has advanced in the differentiation process and is inferred via a diffusion map based manifold learning technique called \emph{dpt} \citep{Haghverdi2016} on 2,000 highly variable genes \citep{Wolf2018,Bergen2020}.
We take measured gene expression levels over pseudotime $t$ as our observations $\Xtrue(t)$. 

In this setting, domain knowledge asserts that GATA genes (\emph{GATA1}, \emph{GATA2}) regulate the expression of hemoglobin subunits (\emph{HBB}, \emph{HBA1}, \emph{HBA2}, \emph{HBD}, \emph{HBQ1}) \citep{Ding2010,Johnson2002,Suzuki2013,Shearstone2016}.
The normalized expression of genes related to these subunits over pseudotime is presented in Figure~\ref{fig:GRN_human} (middle row).
The expressions are scaled between 0 and 1 for each gene before training. 
We first apply C-NODE to these 7 genes with known ground truth.
The bottom row of Figure~\ref{fig:GRN_human} shows the row-wise normalized absolute values of the adjacency matrix inferred by C-NODE.
Our approach properly assigns hemoglobin subunit changes to GATA genes, even though visually the hemoglobin target genes appear to be more correlated among themselves than with the GATA drivers.

We expect the regulatory elements and their target genes to be similar across species.
To show that the previous results are stable across species, we next apply C-NODE to mouse single-cell RNA-seq data from~\citep{Pijuan-Sala2019} (GEO accession number: GSE87038) where blood progenitors similarly differentiate into Erythroid cells.
Consistent with previous results, we also observe in Figure~\ref{fig:supp:GRN_mouse} that hemoglobin genes depend on GATA genes in Erythroid lineage (more details are discussed in Appendix~\ref{app:grn}).

\begin{figure}[t]
\vspace{-1cm}
\centering
\includegraphics[width=0.8\columnwidth]{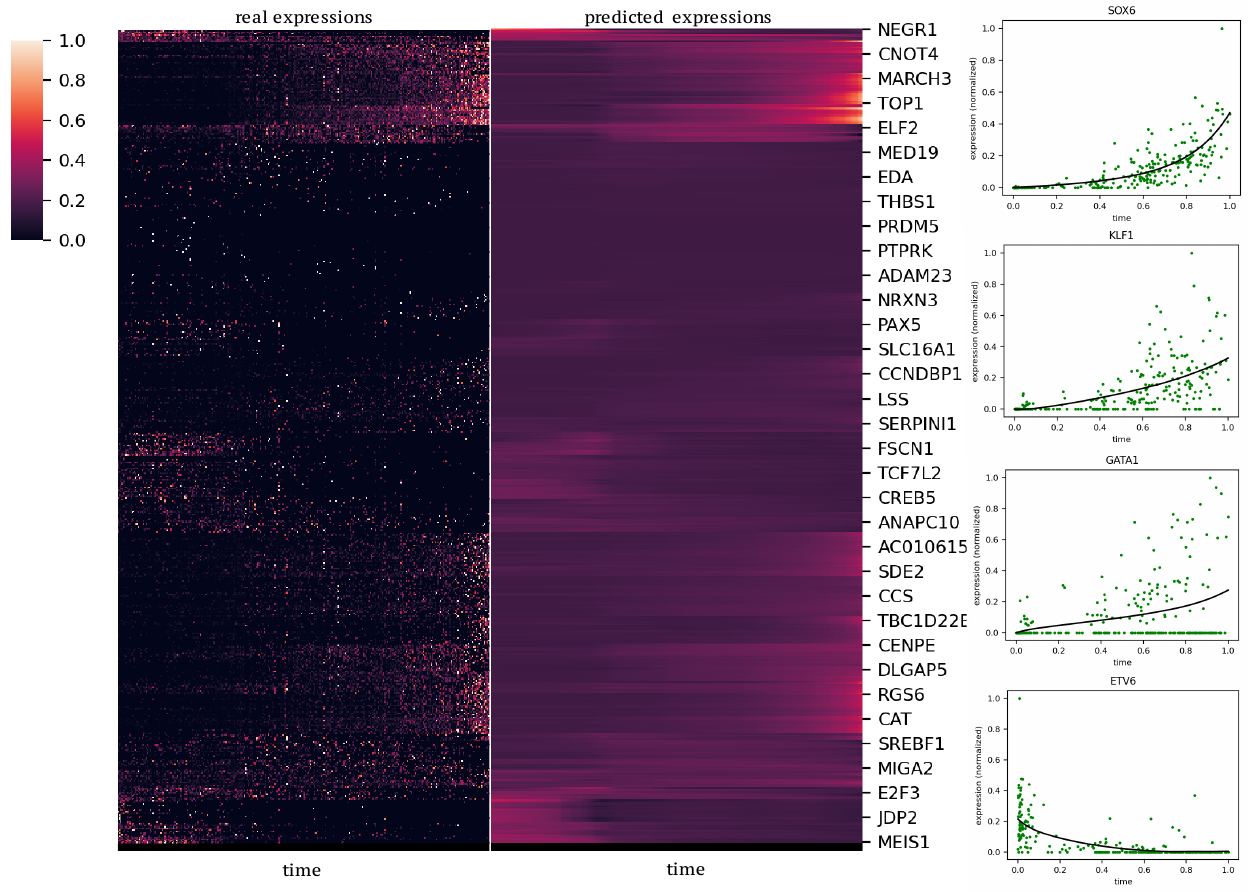}
\caption{Prediction performance using human bone marrow data with 529 genes. Shown are the predictions for 141 important features (on left), and selected transcription factors (on right).}\label{fig:GRN_large}
\vspace{-0.3cm}
\end{figure}

Finally to study the scalability of the proposed method and the effectiveness of the sparsity regularizer, we select 529 highly-variable genes from the whole human hematopoiesis data~\cite{luecken2021a} and apply C-NODE on cells from the Erythroid lineage.
Many of those genes are spurious for the Erythroid lineage as their expression does not change along the lineage.
After training, the model selects 141 key genes as important with at least one interaction to other genes.
Our first observation is that the important genes are ranked as highly variable only for the Erythroid lineage (Figure~\ref{fig:supp:vann}) which shows that C-NODE avoids using spurious features for prediction (Figure~\ref{fig:GRN_large}).
Using the chromatin accessibility features available in the human immune cells datasets as reference, we also observe 22 regulatory genes known as transcription factors among those important genes. 
Assessed by the literature, all the 22 transcription factors are important for the differentiation of the Erythroid lineage.
The expression and the prediction of four transcription factors are shown in Figure~\ref{fig:GRN_large}, left.
Other transcription factors are listed in Table~\ref{table:supp:TFs}.

Since ground truth dynamics are not known for GRNs, validating the inferred interactions is challenging.
In order to assess the biological relevance of the learned interactions, we perform Gene Set Enrichment Analysis (GSEA) via the enrichr method~\cite{Chen2013} (discussed in Appendix~\ref{app:grn}).
The enrichment test also captures many relevant processes including hematopoietic stem cell differentiation, Erythrocyte differentiation, and some immune related signaling pathways. 
This provides evidence that indeed key regulatory processes for the differentiation have been captured by our model. 

\section{Conclusion}
\label{sec:conclusion}
We proposed C-NODE, an approach to identification and causal structure learning of (sparse) ODE-based dynamical systems using Neural ODEs.
First, we observe that even if a method achieves perfect predictive accuracy, it may not be able to predict the system's behavior under interventions, as ODE identification is generally ill-posed.
Therefore, a key focus of our work lies on the predominantly neglected issue of restricting the search space in meaningful ways to recover the underlying sparse system and causal structure.

We devised a simple and practical method to extract causal dependencies (including cyclic relationships) from a learned neural ODE derivative network.
We then demonstrated that C-NODE performs well on causal structure identification for a wide variety of settings and further corroborated our findings by correctly predicting the effect of different forms of interventions targeting both the evolving variables as well as parameters of the governing ODE itself.

In our experiments we analyze C-NODE on synthetic and real-world gene regulatory data with varying numbers of variables, noise levels, and irregular sampling intervals.
While unidentifiability indeed affects vanilla NODE, C-NODE still reliably infers sparse causal structures.
Going forward, our results suggest an in-depth analysis of the conditions under which unidentifiability manifests itself in practice as a fruitful direction for future work.
At the same time, extending C-NODE for successful hypotheses generation in high-dimensional real datasets with stochasticity, delay, unobserved confounding, or heterogeneous environments is an exciting challenge for further research.

Given these limitations, we highlight that caution must be taken when informing consequential decisions, e.g., in healthcare, based on causal structures learned purely from observational data.
At the same time, we hope that causal modelling of dynamical systems broadly and C-NODE in particular can be valuable tools for hypothesis-generation in various scientific domains to suggest promising experiments for in-depth follow up.



\bibliography{main}
\bibliographystyle{tmlr}

\newpage
\appendix
\onecolumn
\section{Discretization of ODE Systems}
\label{app:disc}

The choice to compare primarily to \dynotears{} is motivated by the fact that the autoregressive model in eq.~\eqref{eq:granger} can be viewed as a finite difference approximation to linear systems of ODEs that also incorporates sparsity.
Since this recent method by \citet{Pamfil2020} demonstrates superior performance to most other competing methods for inferring causality from time series data, we chose it as a strong competitor to our method.
For numerical treatment of derivatives in ODEs we often employ \emph{finite differences}, where the derivative at time step $t$ is approximated via differences of function values at slightly different time steps.
For example, the backward finite difference for the first derivative is given by $(f(t) - f(t-h)) / h$ for some $h > 0$.
For time series data, we often use (without loss of generality) $h = 1$ and can thus approximate the derivative via $f(t) - f(t-1)$.
Similar approximations of higher order derivatives require more terms.
Generally, to approximate the $k$-th derivative, information from $k+1$ different time points is needed.
Hence, finite combinations of the form 
\begin{equation}
  \sum_{\tau=0}^{k}W_\tau f(t-\tau)\:,
\end{equation}
which is also used by \dynotears{}, can in principle encode (linear combinations of) derivatives of $f$ at time $t$ up to order $k-1$.
Hence, time-series based methods such as \dynotears{} could in principle be expected to be able to model ODE systems correctly.

\section{Proofs}
\label{app:proofs}

Here, we provide the example of unidentifiability mentioned in the main text.
We start with some notation and known results.
The space of general solutions of $\dot{X} = A X$ forms an $n$-dimensional sub vector space $\cL$ of all continuously differentiable functions from $\bR$ to $\bR^n$.\footnote{The statements in this paragraph are proven in most textbooks on ODEs, e.g., see \citep{hirsch1974differential}.}
For a basis $\phi_1, \ldots, \phi_n$ of $\cL$, we call $\Phi_A = (\phi_1, \ldots, \phi_n) : \bR \to \bR^{n\times n}$ a \emph{fundamental system} of the differential equation, which in this setting is simply given by $\Phi_A(t) = e^{At}$.
The solution to the IVP with $X(t_0) = x_0 \in \bR^n$ is then $X(t) = \Phi_A(t) h$ for an $h$ such that $x_0 = \Phi_A(t_0) h$, which exists since $\Phi_A(t)$ has full rank for all $t$.
Hence, our main goal restricted to $\cFlin$ reads: \emph{Assuming $f^{\star}(X, t) = A^{\star} X$ and w.l.o.g.\ $a = 0$, can we uniquely identify $A^{\star}$ given $\Xtrue$?}
In other words, does $e^{At} x_0 = e^{A't} x_0$ imply $A = A'$?
This is not the case generally as we show with the following example.
Non-uniqueness of $A$ amounts to the existence of $B \in \bR^{n \times n}$ different from $A$ such that $\Phi_A h = \Phi_B g$ for some $g, h \in \bR^n$ with $\Phi_A(0) h = \Phi_B(0) g = X(0)$.
For $n=2$, $X(0) := (1, 1)$ and
\begin{equation}\label{eq:examples}
 A = \begin{pmatrix} 1 & 0 \\ 0 & 1 \end{pmatrix}, \qquad
 B = \begin{pmatrix} 0 & 1 \\ 1 & 0 \end{pmatrix}
\end{equation}
we have $X(t) = e^{At} h = e^{Bt} g = (e^t, e^t)$ for all $t \in \bR$. 

\xhdr{Remarks}
First, we note that this result appears intuitive when writing out the two systems, showing that we can choose any constant initial value $X(0) = (\alpha, \alpha)$.
Another way to understand unidentifiability intuitively for autonomous systems is to consider the flow vector field $f(X)$ of the ODE in which a particular solution is a single one-dimensional trajectory $X(t)\in\bR^n$ following the vector field $f(X)$ at every point.
An alternative system $f'$ also solved by $X(t)$ only needs to preserve the vector field $f(X)$ along its trajectory ($f'|_{X(t)} = f|_{X(t)}$ for all $t \in [a,b]$), but could arbitrarily differ (without violating continuity assumptions) from $f(X)$ outside of $\{X(t) \given t \in [a,b]\}$.

Finally, most results discussed here also hold for inhomogeneous, autonomous, linear systems ($f(X) = AX + b$ for $b \in \bR^n$).
In this case, the $n$-dimensional solution vector space is an affine subspace $\cL + \tilde{X}$, where $\cL$ is the solution vector space of the homogeneous system and $\tilde{X}$ is any specific solution of the inhomogeneous one.

\xhdr{Ill-posedness despite regularization}
Here we show that our main goal is still ill-posed even under sensible sparsity regularizations.
We begin again with the example from eq.~\eqref{eq:examples}, where the matrices $A$ and $B$ have equal values for $\reg{\cdot}$, $\|A\|_1$, and $\|A\|_{1,1}$.
Let $C \in \bR^{2\times 2}$ be any system that has $X(t) = (e^t, e^t)$ as a solution for the initial value $X(0) = (1, 1)$ on all of $\bR$.
Since $X_1 = X_2 = \dot{X}_1 = \dot{X}_2$ on $\bR$ the coefficients in each row of $C$ must sum up to $1$.
Hence, the minimum achievable value for $\|C \|_{1,1} = \sum_{i,j=1}^2 |C_{ij}|$ is 2, which is achieved by both $A$ and $B$.
Similarly, the minimum achievable value for $\|C \|_{1} = \max_{j=1,2} \sum_{i=1}^2 |C_{ij}| = \max\{|C_{11}| + |C_{21}|, |C_{12}| + |C_{22}|\}$ under the row-unit-sum constraint is 1, which is also achieved by both $A$ and $B$.
Finally, in the linear case the minimum achievable $\reg{C} = \approxreg[=0]{C} = \sum_{i,j=1}^2 \B{1}\{C_{ij} \ne 0\}$ is also 1, again achieved by both $A$ and $B$.
Therefore, the systems $A$ and $B$ in eq.~\eqref{eq:examples} are indeed among the minimum complexity solutions under all considered regularization schemes.

While other types of regularization may yield unique solutions (Tikhonov regularization for ill-posed problems), these typically clash with the explicit demand for sparsity in the system, with many relationships not just being weak, but desired to be non-existent.
Our focus lies on sparsity enforcing regularization motivated by $\reg{\cdot}$ throughout.
We also remark the close resemblance of our arguments to the fact that Ridge regularization yields unique solutions in linear regression whereas Lasso may not.
In our example, consider $\|\cdot\|_{2}$ or $\|\cdot\|_{2,2}$ instead.
In this case the minimum complexity system is \emph{uniquely} given by $C_{ij} = 0.5$ for $i,j \in \{1, 2\}$.
It is important to recognize though that this argument does \emph{not} suffice to prove that our statement does not hold for these regularizers.
It merely illustrates that our examples and proof techniques fail for these regularizers.
Refining these unidentifiability results is an interesting direction for future work.

\section{Implementation Details}
\label{app:implementation}

\subsection{Synthetic Data Generation}

We generate random datasets with 10, 20, and 50 variables and at most 200 observations. 
For each dataset, we first start with a random ground truth adjacency matrix and generate the discrete observations using off-the-shelf explicit numerical Runge-Kutta style ODE solvers~\citep{scipy}.
We then add zero-mean, fixed-variance Gaussian noise to each variable and observation independently.
Finally, a percentage of samples (denoted by `irr') is randomly dropped from the dataset to simulate irregular observation times.

\subsection{Architecture and Training Procedure}

All neural networks used in this work are fully connected, feed-forward neural networks.
The initial velocities are predicted using a neural network with two hidden layers with 20 neurons each and $\tanh$ activations.
The main architecture to infer velocities (or accelerations) also contains two hidden layers of sizes 20, 50, or 100 depending on the size of the input and ELU (or for some experiments linear) activation function.
ELU and $\tanh$ were used because they allow for negative values in the ODE~\citep{Norcliffe2020}.
As an ODE solver, we use an explicit $5$-th order Dormand-Prince solver commonly denoted by \texttt{dopri5}.

All models are optimized using Adam~\citep{kingma2017adam} with an initial learning rate of $0.01$.
We use PyTorch's default weight initialization scheme for the weights and set the regularization parameter $\lambda$ for the L1 penalty to $0.01$ in the 10, and 20-dimensional examples and to $0.1$ for the 50-dimensional example.
All models can be trained entirely on CPUs on consumer grade Laptop machines within minutes or hours.
To compute the MSE uncertainty, the experiments in Figures~\ref{fig:linear},\ref{fig:spiral} and \ref{fig:LV} are run 10 times with random seeds.
Finally, the presence of edges in the weight matrices was determined by thresholding absolute values at $0.05$ for the synthetic datasets and $0.001$ for the real dataset.

\section{Extensions}
\label{app:extensions}

\subsection{Measurement noise}

Considering a deterministic version of dynamical systems with measurement noise, we have:
\begin{equation}
    \Tilde{X}(t) = X(t) + E(t).
\end{equation}
Where, $X(t)$ is assumed to be governed by the ODE system and the noises $E_i$ are assumed to be jointly independent with zero mean. 
We show that our causal inference model based on NODEs can be relatively resistant to measurement noise (see the example in Figure~\ref{fig:spiral}). 
However, we can still extend our approach to latent-variable models for more complicated systems.
In a general form, consider a (generative) model $e$ which encodes the initial position $\Tilde{X_0}$ into the latent variable $Z_0$ as follows:
\begin{equation}
    \mathbf{Z_0}=e(\mathbf{\Tilde{X}},\theta_e).
\end{equation}
This is then used by the ODE function $f$ and the ODE solver consequently:
\begin{equation}
    \begin{array}{c}
        \dot{Z} = f(Z,t,\theta)\\
    Z_0,\ldots ,Z_N = ODESolver(f,\mathbf{Z_0}, (t_0,\ldots , t_N))
    \end{array}
\end{equation}
The latent variables are then decoded as follows:
\begin{equation}
    \hat{X}_0,\ldots ,\hat{X}_N = d(Z,\theta_d).  
\end{equation}

Such a latent-variable model is causally interpretable and can be integrated into our method, if we can learn $\dot{X}$ as a function of $\dot{Z}$ naturally from the model.
While most of the extensions to NODEs for noisy and irregularly-sampled observations perform well with respect to the reconstruction accuracy, they are hardly interpretable due to their model complexity~\citep{Rubanova2019, norcliffe2021neural}.

\subsection{Data from Heterogeneous Systems}

Our algorithm could potentially also be extended to (noisy) observations generated from heterogeneous experiments. 
Following \citet{pfister2019learning}, in this case we treat the observations from each experiments as a time-series sample. 
Training the model with heterogeneous samples, we may hope to identify the causal model that is invariant across the experiments \citep{pfister2019learning}.
It is an interesting direction for future work to determine whether heterogeneous experiments allow us to overcome the unidentifiability results in Section~\ref{sec:theory}.

\section{Extended Results}

\subsection{Synthetic Linear ODEs}
\label{app:synthetic}

\begin{figure}
\centering
\includegraphics[width=0.8\textwidth]{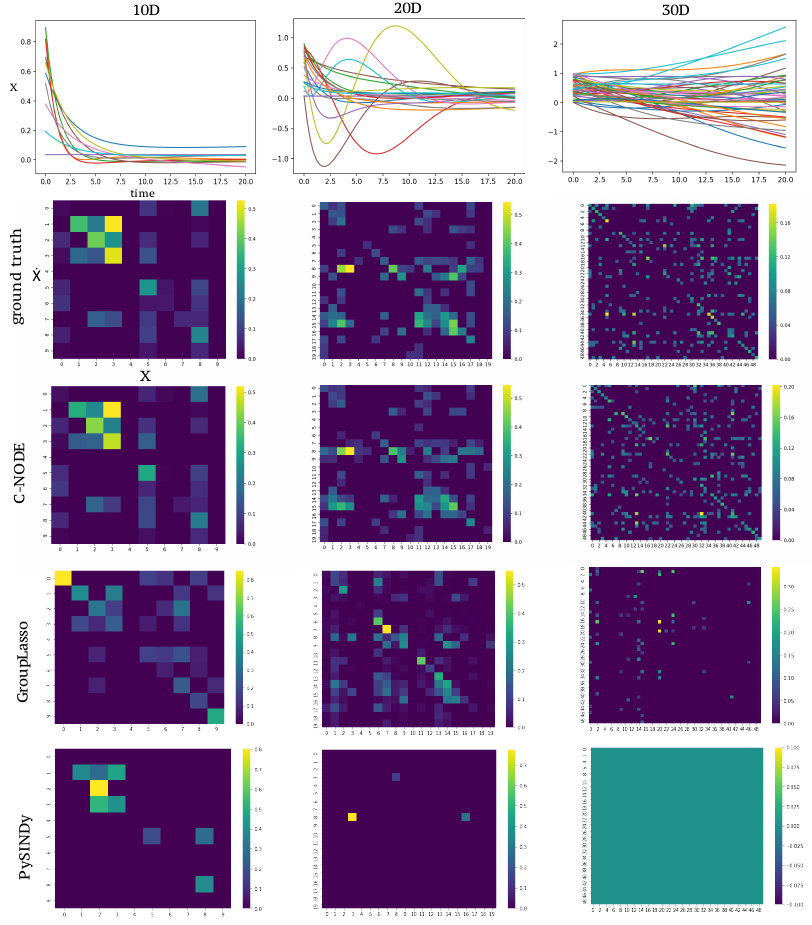}
\vspace{-3mm}
\caption{The synthetic datasets and the inferred adjacency matrices using C-NODE, GroupLasso~\citep{bellot2021graphical} and PySINDy~\citep{desilva2020}. Shown are the absolute values for the adjacency matrices referred to the regularly sampled ($\mathrm{irr}=0.0$) and non-noisy ($\sigma=0$) setting.
}\label{fig:synthetic:adjacency}
\end{figure}

\begin{figure}[ht]
\centering
\includegraphics[width=0.8\textwidth]{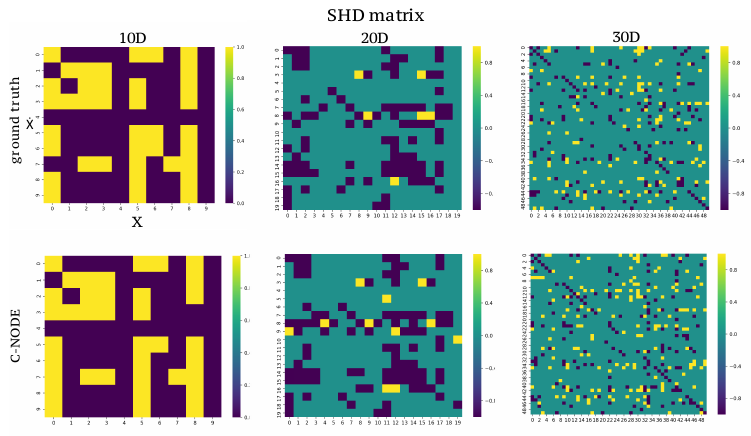}
\caption{The true and the inferred $\overline{\text{SHD}}$ matrices for the synthetic datasets using C-NODE, related to Table~\ref{tab:results}. }
\label{fig:SHD}
\end{figure}

The learned adjacency matrices using vanilla NODE~\cite{chen2018neural}, GroupLasso~\citep{bellot2021graphical} and PySINDy~\citep{desilva2020} are illustrated in Figure~\ref{fig:synthetic:adjacency}.

The SHD matrices using C-NODe are presented in Figure~\ref{fig:SHD}.
$S_{ij}$ in an SHD matrix represents the presence or absence of an edge between $X_i$ and $X_j$ including the sign of the effect (e.g., $S_{ij}=-1$ means $X_j\longrightarrow X_i$ with a negative coefficient). 

\subsection{Chemical reaction networks}
\label{app:chemical}

\begin{figure}[ht]
\centering
\includegraphics[width=0.7\textwidth]{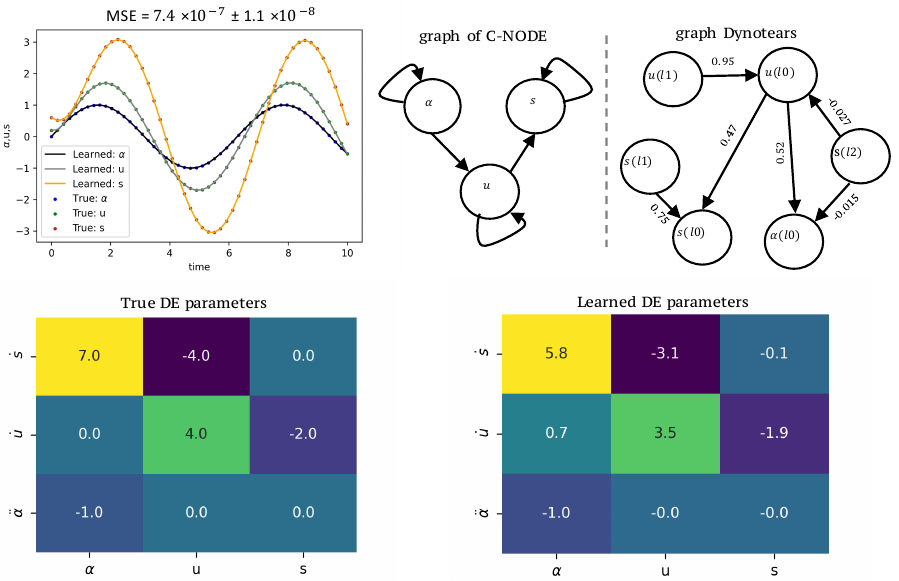}
\caption{Example of a chemical reaction network modeling the transcriptional dynamics of a gene.}\label{fig:chemical}
\end{figure}

Here, we study a model of transcriptional dynamics which captures transcriptional induction and repression of unspliced precursor mRNAs $u(t)$ splicing into mature mRNAs $s(t)$ at rate $\beta$.
The mature mRNAs eventually degrade with rate $\gamma$. 
\begin{equation}
  u =  \alpha - \beta u, \qquad
  s =  \beta u - \gamma s,
\end{equation}
where $\alpha$ is the reaction rate of the transcription. 
We assume $\beta$ and $\gamma$ to be constant and the transcription rate $\alpha$ to vary over time. 
The results in Figure~\ref{fig:chemical} show that our method can successfully learn the structural graph as well as the ODE parameters while \dynotears{} fails on the same task.
Note that in this case we add $\alpha$ as a variable in the system with a fixed, pre-specified time-dependence in such a way that it satisfies an ODE separately from $u, s$.
Our method successfully identifies this structure where the evolution of $\alpha$ does not depend on $u$ and $s$, but conversely, the derivatives of $u$ and $s$ depend on $\alpha$.
Encouraged by these synthetic results, we also tested our method on real single-cell gene expression data, described in Section~\ref{sec:experiments}.

\subsection{Gene Regulatory Network Inference}
\label{app:grn}
\begin{figure}
\centering
\includegraphics[width=0.7\columnwidth]{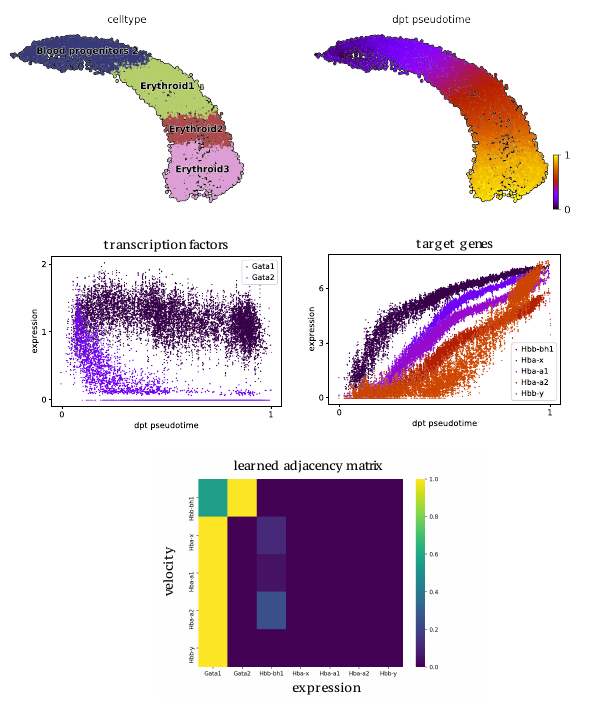}
\caption{Gene regulatory network inference results using mouse dataset  with 7 genes.}\label{fig:supp:GRN_mouse}
\end{figure}

We apply C-NODE to real mouse single-cell RNA-seq data from~\citep{Pijuan-Sala2019} (GEO accession number: GSE87038).
We select a branch of data in which blood progenitors differentiate into Erythroid cells with a total of 9,192 cells (or samples). 
The count matrix is normalized to
one million counts per cell.
We randomly subset 300 cells from all 9,192 as training data for C-NODE. 
We observe in Figure~\ref{fig:supp:GRN_mouse} that the inferred GRN using GATA and hemoglobin genes resembles the one in Figure~\ref{fig:GRN_human}.

\begin{figure}
\centering
\includegraphics[width=1\textwidth]{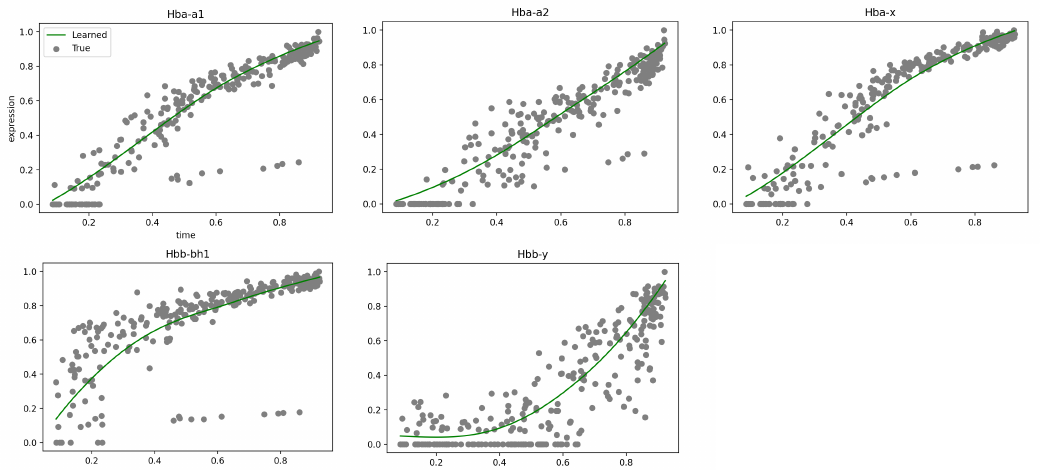}
\caption{Predictions of the expression of the target genes together with the 300 samples used for training, related to Figure~\ref{fig:supp:GRN_mouse}.\label{fig:GRN:pred}}
\end{figure}

\begin{figure}
\centering
\includegraphics[width=1\textwidth]{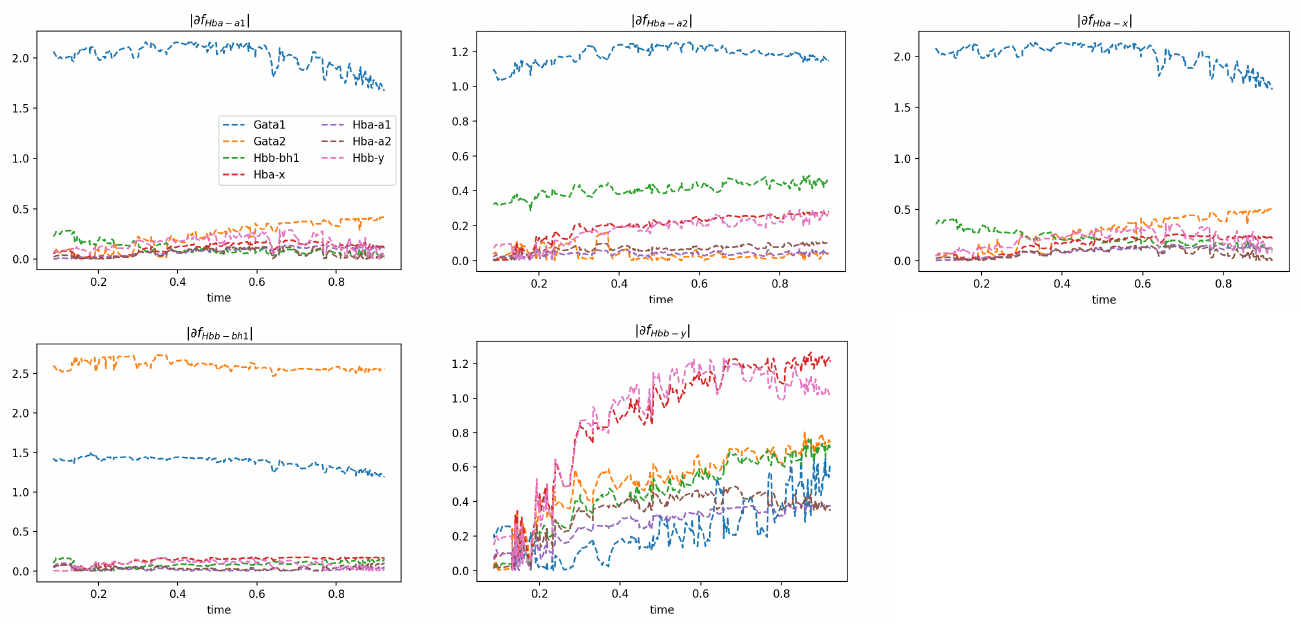}
\caption{Gradients for the target genes, related to Figure~\ref{fig:supp:GRN_mouse}.\label{fig:GRN:sens}}
\end{figure}

In Figure~\ref{fig:GRN:pred} and Figure~\ref{fig:GRN:sens}, we show additional results for GRN inference of single-cell mouse data.
Figure~\ref{fig:GRN:pred} shows the predictions of C-NODE.
The partial derivatives in Figure~\ref{fig:GRN:sens} indicate that despite non-linear activations, the associations are mostly linear for the target genes except for \emph{Hbb-y}. 
This suggests that linear $f^{\star} \in \cFlin$ may indeed be a decent approximation for certain gene regulatory networks.

For the large dataset shown in Figure~\ref{fig:GRN:pred}, the gene set enrichment analysis uses a priori gene sets that involve in known biological pathways. 
For each list of dependent genes inferred using C-NODE, we then analyze whether the majority of genes in each pathway fall in the extremes of this list.

\begin{figure}
\centering
\includegraphics[width=0.5\columnwidth]{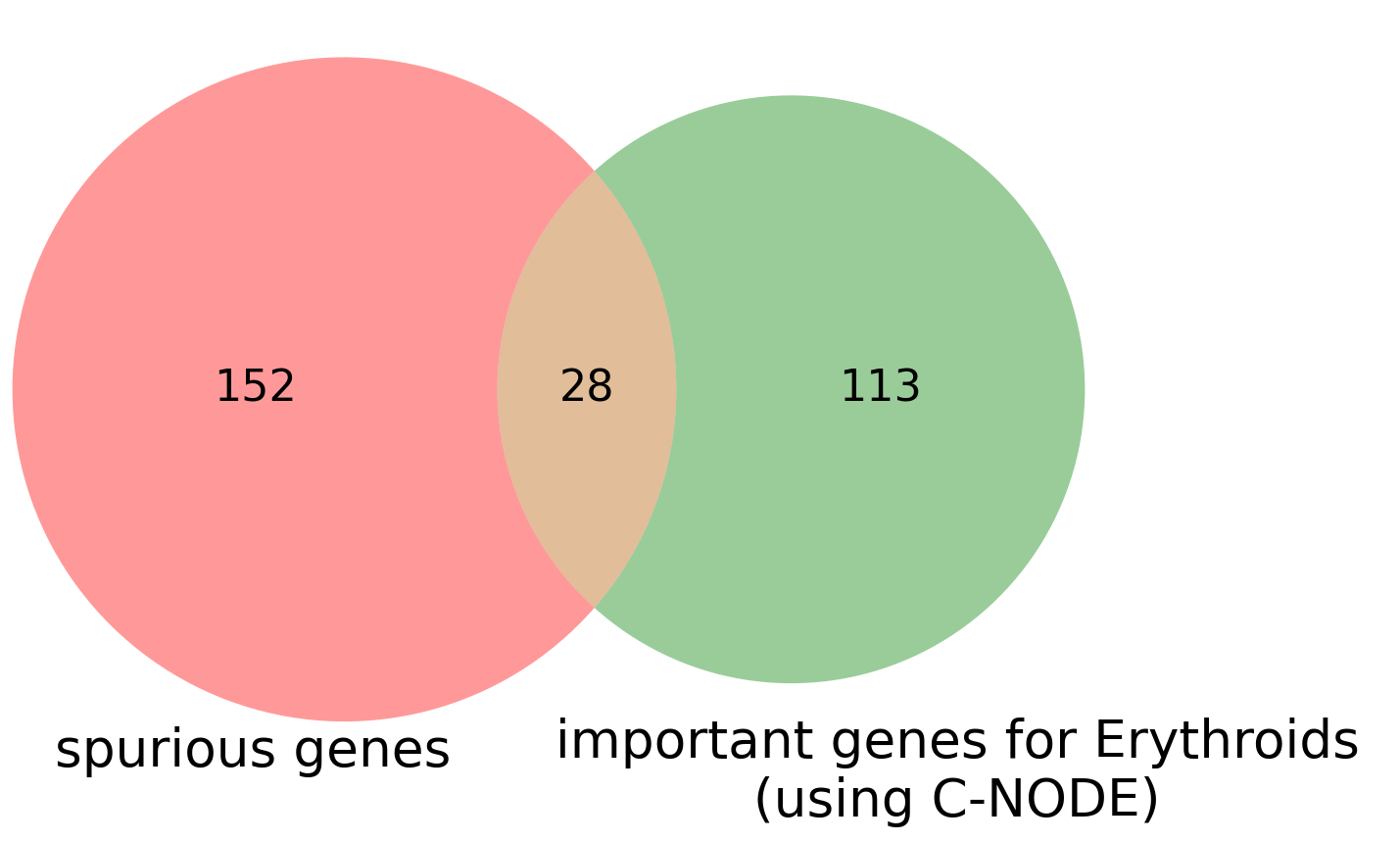}
\caption{The important genes extracted using C-NODE are Erythroid specific and have negligible overlap with spurious genes.}\label{fig:supp:vann}
\end{figure}

\begin{table}[t] 
\caption{The inferred TFs for Erythroids using C-NODE.}\label{table:supp:TFs}
\begin{center}
\begin{tabular}{l}
\toprule
 CTCF, E2F1, E2F3, E2F8, ETV6, GATA1, GFI1B, KLF1,\\
 KLF13, MAFG, MAZ, MXI1, MYBL2, NFE2L2, NFIA, SP4,\\ 
 RREB1,RUNX1, SOX4, SOX6, SREBF1, TAL1\\
 \bottomrule
\end{tabular}

\end{center}
\end{table}

\end{document}